\pdfoutput=1
\documentclass[11pt]{article}

\usepackage[preprint]{acl}

\usepackage{times}
\usepackage{latexsym}
\usepackage[T1]{fontenc}
\usepackage[utf8]{inputenc}
\usepackage{microtype}
\usepackage{inconsolata}
\usepackage{graphicx}
\usepackage{tcolorbox}
\usepackage{booktabs}
\usepackage{multirow}
\usepackage{makecell}

\title{(Fact) Check Your Bias}

\author{Eivind Morris Bakke \\
  University of Oslo \\
  Oslo, Norway \\
  \texttt{eivindmb@ifi.uio.no} \\\And
  Nora Winger Heggelund \\
  University of Oslo \\
Oslo, Norway \\
  \texttt{norawh@ifi.uio.no} \\}

\begin{document}
\maketitle
\begin{abstract}

Automatic fact verification systems increasingly rely on large language models (LLMs). We investigate how parametric knowledge biases in these models affect fact-checking outcomes of the HerO system (baseline for FEVER-25). We examine how the system is affected by: (1) potential bias in Llama 3.1's parametric knowledge and (2) intentionally injected bias. When prompted directly to perform fact-verification, Llama 3.1 labels nearly half the claims as "Not Enough Evidence". Using only its parametric knowledge it is able to reach a verdict on the remaining half of the claims. In the second experiment, we prompt the model to generate supporting, refuting, or neutral fact-checking documents. These prompts significantly influence retrieval outcomes, with approximately 50\% of retrieved evidence being unique to each perspective. Notably, the model sometimes refuses to generate supporting documents for claims it believes to be false, creating an inherent negative bias. Despite differences in retrieved evidence, final verdict predictions show stability across prompting strategies. The code is available at: \url{https://github.com/eibakke/FEVER-8-Shared-Task}

\end{abstract}

\section{Introduction}
In modern society, the rapid spread of information creates significant opportunities for misinformation. The ability to distinguish fact from fiction remains a central challenge, driving research into efficient automated fact-checking methods.

Our work builds on the HerO system \cite{yoon-etal-2024-hero} which serves as the baseline for the 2025 FEVER Workshop. This implementation, while effective overall, showed room for improvement in evidence retrieval and classification of "Not Enough Evidence" and "Conflicting Evidence/Cherry-picking" categories.

Given HerO's reliance on LLM document generation in the initial retrieval pipeline, and the known tendency of LLMs to exhibit bias from their parametric knowledge, we hypothesized that LLM bias may be a part of the reason for the HerO system's performance. To study this effect we investigate two central hypotheses: \begin{enumerate}
    \item \textbf{LLM-inherent bias hypothesis}: The LLM generating hypothetical fact-checking documents in the HerO-system contains biases in the parametric knowledge. 
    \item \textbf{Bias propagation hypothesis}: These biases systematically affect downstream components of the fact-checking pipeline, specifically evidence retrieval and final veracity prediction.
\end{enumerate}
To test these hypotheses, we conduct two experiments: first we examine how the HerO-system performs without external knowledge, relying solely on the parametric knowledge of the LLM; second, we investigate how intentionally prompted biases in hypothetical document generation affect evidence retrieval and verification decisions. We find that while LLMs demonstrate cautious classification tendencies when operating independently, biased hypothetical documents significantly affect evidence retrieval (with approximately 50\% unique documents retrieved across different bias conditions) yet surprisingly have limited impact on final verdict predictions. We also discover that under certain conditions, the Llama 3.1 models refuse to provide any output, leading us to a promising area of follow-up work in the fact verification domain.

\subsection{Terminology}
The term bias encompasses various meanings, including statistical biases (e.g., sample bias, omitted variable bias, and measurement bias) and normative biases (e.g., those that lead to unfair or unequal outcomes, often due to human biases reflected in training data) \cite{Olteanu2019, Campolo2017}. The latter may involve differential treatment by the model, for example, based on gender, religion, culture, or political alignment. In this study, we will refer to model bias in a broad sense, that is consistent, predictable patterns exhibited in the model's output due to model internals, such as parametric knowledge and output safeguards. Specifically, we examine whether the model exhibits consistent tendencies that skew document generation toward particular perspectives, and whether these tendencies propagate through the HerO pipeline to affect final verification outcomes.

\section{Related work}
\subsection{Fact-checking}
Vlachos and Riedel presented how the fact-checking process consists of different stages, each of which may be automated \cite{vlachos-riedel-2014-fact}. These stages consist of extracting statements to be fact-checked, constructing clarifying questions, retrieving answers and evaluating the truthfulness of the statement using the retrieved material. Several LLM based systems for automatic fact-checking have been developed in recent years with widely different architectures, including fine-tuning LLMs to evaluate truthfulness \cite{Choi2024}, knowledge-graphs \cite{kim-choi-2020-unsupervised} and different RAG-implementations \cite{Hai2025}. There have also been attempts to use the parametric knowledge of a language model to perform fact-checking \cite{hoes_altay_bermeo_2023}.

In this paper, we investigate the Herd of Open LLMs for verifying real-world claims (HerO) fact-checking system \cite{yoon-etal-2024-hero}. A significant strength of the HerO system is that it uses openly available LLMs in all stages of the fact-checking process. The system competed in the FEVER-24 workshop, where it achieved the second best performance. Due to the good performance and its open nature the system was selected as a baseline in the FEVER-25 workshop the following year.

\subsection{Knowledge conflicts and bias in LLMs}
LLMs in fact-checking systems may suffer from multiple shortcomings: they may reach wrong conclusions due to conflicting knowledge, generate unsupported answers or propagate biases from training data. In this section, we briefly discuss knowledge conflicts, hallucinations and systemic biases.

\paragraph{Knowledge Conflicts}
Xu et al. describe how conflicts may arise if there are discrepancies between context (user prompt, dialog history and retrieved documents) and parametric knowledge of the model \cite{xu-etal-2024-knowledge-conflicts}. In addition, there may be conflicting information internally in both the context and in the parametric knowledge (inter-context conflict and intra-memory conflict) \cite{xu-etal-2024-knowledge-conflicts}. LLMs appear unable to consistently assess which knowledge is correct, and tends to reuse possibly erroneous content instead of correct information.

\paragraph{RAG Hallucinations}
Retrieval-Augmented Generation (RAG) architectures aim to reduce hallucinations by combining a retrieval module, which identifies relevant pieces of information in a knowledge base, with a generation module, where a LLM uses the retrieved information to produce grounded answers. Béchard and Ayala demonstrated that RAG systems hallucinate less than a fine-tuned LLM on its own \cite{Ayala_2024}. Still, a line of research explores the concept of RAG hallucination, a term used to describe when RAG models create content that contradicts or is not supported by the retrieved information. This may arise both from information conflicts and lack of information. An evaluation of commercial RAG systems for legal texts in the US showed that between 17 and 33 percent of the queries resulted in answers that contained a hallucination \cite{magesh2024hallucinationfreeassessingreliabilityleading}. Sun et al. investigate mechanisms causing hallucinations in RAG systems. They find that a central cause to hallucinations is insufficient utilization of external context and over-reliance on parametric knowledge \cite{sun2025redeepdetectinghallucinationretrievalaugmented}.  

\paragraph{Systemic biases}
Retrieval systems may also be vulnerable to biases resulting from biased training data. Lin et al. find that false information generated by LLMs tend to be replications of popular misconceptions \cite{lin-etal-2022-truthfulqa}. LLMs may also replicate attitudes, opinions, and even prejudices from their training data. Several recent studies have demonstrated biases in LLMs, like leanings toward certain political opinions and parties \cite{rettenberger2025llm_bias}, religious bias \cite{abid2021persistentantimuslimbiaslarge} and gender bias \cite{zhao2024genderbiaslargelanguage, soundararajan-delany-2024-investigating, beatty2024revealinghiddenbiasai}.

\subsection{The HerO system for fact verification}
The HerO system \cite{yoon-etal-2024-hero}, which achieved second place in the FEVER-24 challenge, serves as the baseline for the 2025 FEVER Workshop. It employs a three-stage pipeline using publicly available LLMs: (1) evidence retrieval using Llama 3.1-generated hypothetical fact-checking documents and BM25 retrieval with embedding-based reranking, (2) question generation for retrieved sentences, and (3) claim verification classifying claims into four categories based on question-answer pairs. Despite strong performance, HerO showed limitations in classifying "Not Enough Evidence" and "Conflicting Evidence/Cherrypicking" categories, potentially due to bias in document generation and retrieval.

\subsection{Research gap and our contribution}
Our work bridges fact-checking methodology and LLM bias research by investigating how parametric knowledge in Llama 3.1, and potential inherent biases, impact fact verification. While existing literature has examined knowledge conflicts \cite{xu-etal-2024-knowledge-conflicts}, hallucinations in retrieval systems \cite{sun2025redeepdetectinghallucinationretrievalaugmented, magesh2024hallucinationfreeassessingreliabilityleading}, and systemic biases \cite{rettenberger2025llm_bias, abid2021persistentantimuslimbiaslarge}, our research contributes by: (1) quantifying how LLM-inherent biases affect evidence retrieval in a practical fact-checking system, (2) demonstrating the surprising stability of verdict predictions despite significant retrieval differences, and (3) revealing asymmetric safeguarding behaviors that create inherent negative bias in verification systems. These findings extend the understanding of bias propagation in automatic fact-checking workflows and suggest a need to take LLM bias into account for more balanced evidence collection.

\section{Methods}

Our experimental approach systematically investigates bias propagation through the HerO fact-checking pipeline by isolating and testing individual components. We designed two complementary experiments to: (1) characterize bias due to parametric knowledge in Llama 3.1 through direct prediction without external knowledge, (2) quantify how biased hypothetical fact-checking documents affect evidence retrieval and veracity prediction. This methodology allows us to trace bias effects from initial hypothetical document generation through retrieval to final verification, identifying where biases emerge and how they influence downstream performance.

\subsection{Our baseline - an adapted HerO system}
We used the smaller 8B model versions of Llama 3.1 for all steps in the HerO pipeline, whereas the original used 70B models for both hypothetical fact-checking document generation and veracity prediction. Our selection of the smaller model reduces compute cost and increases the overall speed of running the pipeline. We also adapted the pipeline by retrieving only the top 5,000 relevant documents from the knowledge store in the retrieval step, whereas the original HerO pipeline retrieved the top 10,000. These alterations likely impacted the quality of our predictions, but the methods and the relevancy of our findings about bias likely apply also to the larger model sizes, even if the larger models have bias in a somewhat different direction \cite{rettenberger2025llm_bias}.

\subsection{Data}
We conducted our experimentation with the Averitec dataset \cite{schlichtkrull2023averitec} provided for the FEVER workshop challenge. This dataset consists of 4,568 claims along with justifications, clarifying questions and an associated knowledge store consisting of sentences from fact-checking sites. In our work we only had access to the training and the development sets, as the test set was withheld for the task evaluation. Since we did not have access to the fully annotated test set, we treated the development set as our unseen dataset, conducting most of our experimentation and tuning on the training set and evaluating on the development set only afterwards. Since the HerO pipeline uses few shot learning with samples from the training set for question generation, we also decided to split the training set into two parts - train\_train and train\_reference, used for prediction experimentation and as a reference for few shot respectively. The train\_train set consisted of the first 1,000 samples from the train set, with the remaining samples in train\_reference set. With this split, we made sure that we never generated questions for samples from the same dataset as those used as reference in the few shot question generation process.

\subsection{Experiment 1: Evaluating bias with direct veracity prediction}
\begin{figure}[t]
  \includegraphics[width=\columnwidth]{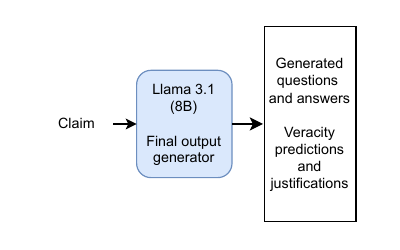}
  \caption{Our direct to prediction system, made to assess the bias due to parametric knowledge inherent in the model. Here a Llama 3.1 8B model is asked to generate output in the final format.}
  \label{fig:direct_system}
\end{figure}

The goal of our first experiment was to evaluate bias due to parametric knowledge in the model used for fact-checking document generation. In the HerO-pipeline, the generated fact-checking documents are subsequently used to retrieve information from the external knowledge store. Consequently, the model's parametric knowledge of the topic at hand will shape the content of the generated document, and by this also influence which external documents are retrieved. This introduces the potential for bias within the retrieval component.

\paragraph{Setup}
To investigate the bias due to parametric knowledge of Llama 3.1, we implemented a direct prediction approach that bypasses all knowledge retrieval components in the in the HerO-system. This simple system design is shown in figure \ref{fig:direct_system}. Our direct prediction system used the Meta-Llama-3.1-8B-Instruct model to classify claims into four categories: "Supported," "Refuted," "Not Enough Evidence," or "Conflicting Evidence/Cherrypicking", and to generate three relevant questions and answers to help verify the claim, using only its internal knowledge. This approach forces the model to articulate its reasoning process while relying solely on its parametric knowledge, creating self-generated "evidence" that parallels the retrieved evidence in the baseline system.

\subsection{Experiment 2: Evaluating the effects of biased hypothetical fact-checking documents}
\begin{figure*}[ht]
    \centering
    \includegraphics[width=.99\linewidth]{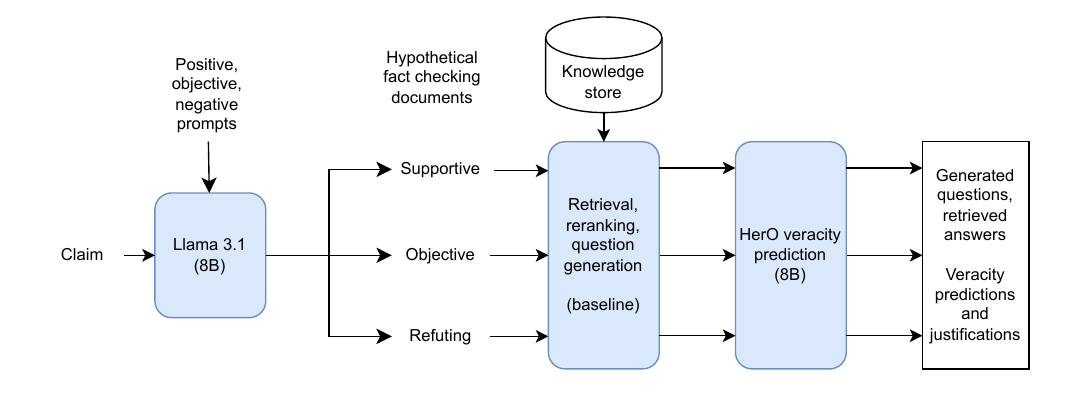}
    \caption{Our multi-prediction system, made to assess the impact of intentionally introduced bias into the retrieval and veracity prediction parts of the system. A modification of the original HerO system, in that we ran three parallel pipelines, with differing biases in their prompts.}
    \label{fig:multi_fc_system}
\end{figure*}

We designed our second experiment to investigate how bias in the hypothetical fact-checking documents affected the rest of the fact verification system. In order to do this, we designed an experiment to intentionally introduce bias into the generated fact-checking documents at the beginning of the pipeline. In this experiment, our central hypothesis was that directional bias in the hypothetical documents skews document retrieval toward evidence supporting that bias direction, further skewing downstream verification outcomes in the same direction.

\paragraph{Setup}
To investigate our hypothesis, we implemented a version of the baseline HerO pipeline which would generate biased documents in the first phase and then run parallel pipelines with the biased documents up to the veracity predictions. Figure \ref{fig:multi_fc_system} shows our pipeline. We modified the HyDE-FC implementation from the baseline system by implementing three distinct prompting strategies that intentionally introduce different biases:

\begin{itemize}
    \item Positive Bias: The model was explicitly instructed to generate a fact-checking passage that supports the claim, highlighting evidence in favor of it. (See figure \ref{fig:hyde-fc-positive-prompt} for the claim and sample document generated.) \
    \item Negative Bias: The model was instructed to generate a fact-checking passage that refutes the claim, highlighting evidence against it. (See figure \ref{fig:hyde-fc-negative-prompt} for the claim and sample document generated.) \
    \item Control: The model was instructed to generate a balanced fact-checking passage, presenting evidence both for and against the claim. (See figure \ref{fig:hyde-fc-objective-prompt} for the claim and sample document generated.)
\end{itemize}

Following the generation of these biased documents, we ran separate, parallel retrieval, reranking, question generation, and veracity prediction processes for each bias condition, identical to the original processes from the baseline. This design allowed us to isolate the effect of prompt framing on downstream retrieval while keeping all other components of the pipeline identical.

\begin{figure}[ht]
\begin{tcolorbox}[colback=white, fontupper=\small]
\textbf{Please write a fact-checking article passage that SUPPORTS the following claim, highlighting evidence in favor of it.}\\
\textbf{Claim}: \textit{In a letter to Steve Jobs, Sean Connery refused to appear in an apple commercial.}\\
\textbf{Passage}: \textcolor{blue}{A 2008 biography by Walter Isaacson, "Steve Jobs," revealed that Sean Connery was initially approached to appear in an Apple commercial for the Apple II computer ...}
\end{tcolorbox}
\centering
\caption{An example of the instruction prompt used for the positively prompted HyDE-FC and its output. The bold text is the instruction, the italic text is a claim, and the blue text indicates the model output.}
\label{fig:hyde-fc-positive-prompt}
\end{figure}

\begin{figure}[ht]
\begin{tcolorbox}[colback=white, fontupper=\small]
\textbf{Please write a fact-checking article passage that REFUTES the following claim, highlighting evidence against it.}\\
\textbf{Claim}: \textit{In a letter to Steve Jobs, Sean Connery refused to appear in an apple commercial.}\\
\textbf{Passage}: \textcolor{blue}{Contrary to a popular myth, Sean Connery, the renowned Scottish actor, never declined an offer to appear in an Apple commercial. ...}
\end{tcolorbox}
\centering
\caption{An example of the instruction prompt used for the negatively prompted HyDE-FC and its output. The bold text is the instruction, the italic text is a claim, and the blue text indicates the model output.}
\label{fig:hyde-fc-negative-prompt}
\end{figure}

\begin{figure}[ht]
\begin{tcolorbox}[colback=white, fontupper=\small]
\textbf{Please write an objective fact-checking article passage about the following claim, presenting a balanced view of evidence both for and against it.}\\
\textbf{Claim}: \textit{In a letter to Steve Jobs, Sean Connery refused to appear in an apple commercial.}\\
\textbf{Passage}: \textcolor{blue}{Fact-Checking the Claim: Sean Connery's Supposed Refusal to Appear in an Apple Commercial The claim that ...}
\end{tcolorbox}
\centering
\caption{An example of the instruction prompt used for the objectively prompted HyDE-FC and its output. The bold text is the instruction, the italic text is a claim, and the blue text indicates the model output.}
\label{fig:hyde-fc-objective-prompt}
\end{figure}

\subsection{Technical description}
We ran the experiments using a GPU with 40 GB standard memory, 24 CPU cores and 250 GB RAM. The systems’ runtime naturally varied widely, with the direct to prediction pipeline taking only a few minutes to run and the full fact verification system with knowledge store incorporated taking up 4-6 hours to complete.

\section{Evaluation methodology and results}

\subsection{Evaluation methodology}

Our evaluation focuses on comparing label distributions, measuring inter-system agreement rates, analyzing semantic similarity of justifications, and quantifying retrieval overlap when relevant through Jaccard similarity and Kendall rank correlation metrics.

\paragraph{Label distribution analysis}
We analyze systematic preferences by comparing label distributions between systems, calculating absolute percentage shifts across the four categories (Supported, Refuted, Not Enough Evidence, Conflicting Evidence/Cherrypicking). This identifies directional bias tendencies.

\paragraph{Inter-system agreement}
For Experiment 1, we measure agreement rates between the direct prediction and the baseline systems across label categories to understand where parametric knowledge aligns with evidence-based decisions.

\paragraph{Semantic similarity analysis of justifications}
We assess semantic similarity of justifications using cosine similarity with all-MiniLM-L6-v2\footnote{The all-MiniLM-L6-v2 sentence transformer is available at https://huggingface.co/sentence-transformers/all-MiniLM-L6-v2, and is a finetuned version of microsoft/MiniLM-L12-H384-uncased which itself was presented in \cite{wang2020minilmdeepselfattentiondistillation}} sentence embeddings across systems, measuring both overall similarity and agreement-conditional similarity.

\paragraph{Retrieval impact analysis}
For Experiment 2, we quantify retrieval differences using: (1) Jaccard similarity between document sets to measure overlap, (2) Kendall rank correlations to assess ranking consistency, and (3) domain distribution analysis to identify source bias patterns.

\paragraph{Performance metrics}
We employ legacy AVeriTeC metrics (question-only, question-answer, and overall scores) to assess whether bias affects final system performance, providing a counterfactual evaluation framework.

\subsection{Results experiment 1: Evaluating bias in hypothetical document generation}

\begin{table}[ht]
\small
\centering
\resizebox{1.05\columnwidth}{!}{
\begin{tabular}{lccc}
\toprule
Label & Direct & Baseline & Absolute Shift \\
\midrule
\makecell{Conflicting Evidence/\\Cherrypicking} & 33 & 12 & -4.20\% \\
Not Enough Evidence & 235 & 12 & -44.60\% \\
Refuted & 133 & 334 & 40.20\% \\
Supported & 99 & 142 & 8.60\% \\
\bottomrule
\end{tabular}
}
\caption{Label distribution shift from direct to knowledge-based baseline prediction}
\label{tab:direct_label_distribution_shift}
\end{table}

The direct and the knowledge-enhanced baseline predictions show remarkably different label distributions (Table \ref{tab:direct_label_distribution_shift}). Agreement between the approaches was low overall (31.20\%), with the highest agreement for "Refuted" claims (81.20\%) and the lowest for "Not Enough Evidence" (3.40\%) and "Conflicting Evidence/Cherrypicking" (6.06\%) verdicts.

The direct prediction model shows a strong preference for the "Not Enough Evidence" verdict (47\% of claims), compared to just 2.40\% for the knowledge-based baseline approach. This pattern reveals a cautious bias in the direct prediction approach (see Appendix A for an example of this behavior). On the other hand, even without the knowledge base, the model still was able to reach a verdict for about the half of the claims by relying solely on its parametric knowledge.

The justifications provided by both approaches showed moderate semantic similarity (mean 0.589, median 0.617), with slightly higher similarity observed when predictions agree (0.606) compared to when they disagree (0.580). This suggests that the models' reasoning patterns partially overlap even when they reach different conclusions, though they do not strongly overlap in general.

\subsection{Results experiment 2: Evaluating the effects of biased hypothetical fact-checking documents}
\label{sec:results_2}

Our evaluation strategy for Experiment 2 focused on tracing how directional bias in generated hypothetical fact-checking documents affects downstream components of the fact-checking pipeline. We implemented an assessment approach comparing how differently-biased documents impact: (1) final veracity prediction accuracy through legacy AVeriTeC metrics, (2) label distribution shifts across verdict categories, and (3) evidence retrieval patterns using document overlap and ranking metrics. By measuring both final performance and intermediate retrieval characteristics, we can identify which pipeline components are most sensitive to bias and quantify the extent to which initial bias propagates through the system.

\paragraph{Retrieval metrics}
Our analysis of the documents retrieved for the 500 claims in the development set revealed substantial differences in retrieval outcomes across the three bias conditions. Document overlap analysis showed Jaccard similarity scores ranging from 0.42 to 0.56, indicating that different biases led to approximately half of the retrieved documents being unique to that perspective compared to the other perspectives. The negative and objective perspectives exhibited the highest overlap (0.56), while positive and negative perspectives were most distinct (0.42). Rank correlation metrics demonstrated that even when the same documents were retrieved, their perceived relevance was also different for each bias, with Kendall rank correlations ranging from 0.48 between positive and negative to 0.61 between negative and objective. Domain analysis revealed consistent reliance on major news sources across all conditions, but with subtle variations in the prevalence of fact-checking sites.

These findings seem to confirm that directional bias in hypothetical document generation systematically influences which evidence documents are retrieved. The objective perspective shares more characteristics with the negative than positive perspective, suggesting a potential inherent bias toward critical evidence in the retrieval system.

\paragraph{Final output evaluation, legacy AVeriTec scores and distributional shifts in labels predicted}

Analysis of label distribution shifts across different prompting strategies (Table \ref{tab:label_distribution_comparison}) reveals surprising stability in final verdicts despite substantial differences in retrieved evidence. The positive-biased system showed the largest shifts (Table \ref{tab:label_distribution_comparison}), with a 4\% absolute decrease in "Refuted" verdicts and corresponding increase in "Supported" verdicts compared to the baseline, aligning with its intended intentionally introduced bias. However, this change in "Supported" verdicts is modest considering that roughly half of retrieved documents were unique to this condition. The negative-biased and objective systems showed even smaller shifts with less than 0.8\% absolute shift across all categories. 

This stability is further reflected in the (legacy) AVeriTeC scores (Table \ref{tab:legacy_averitec_scores}), where all approaches achieved similar performance. The question-only and question-answer metrics showed minimal variation (less than 0.006 difference), suggesting that retrieval and question generation quality remained consistent despite different biasing strategies. Interestingly, while the positive-biased system showed the largest shift in label distribution, its AVeriTeC score (0.48) was only marginally lower than the baseline (0.518). The objective approach slightly outperformed the baseline (0.522 vs 0.518). Overall, these results indicate that the veracity prediction component demonstrates robustness to variations in retrieved evidence. Appendix B. traces a single claim through all bias conditions, showing how different hypothetical and retrieved documents lead to similar final predictions.

\begin{table}[ht]
\small
\centering
\resizebox{\columnwidth}{!}{
\begin{tabular}{lrrrrrrrr}
\toprule
\multirow{2}{*}{Label} & \multicolumn{2}{c}{Baseline} & Positive & Negative & Objective \\
\cmidrule(lr){2-3} 
 & Count & \% & \% shift & \% shift & \% shift \\
\midrule
Supported & 142 & 28.4 & +4.0 & 0.0 & -0.2 \\
Refuted & 334 & 66.8 & -4.0 & -0.8 & 0.0 \\
Not Enough Evidence & 12 & 2.4 & 0.0 & 0.0 & +0.8 \\
\makecell{Conflicting Evidence/\\Cherrypicking} & 12 & 2.4 & 0.0 & +0.8 & -0.6 \\
\bottomrule
\end{tabular}
}
\caption{Label distribution across different prompting strategies. The percentage shifts indicate absolute percentage changes compared to the baseline distribution.}
\label{tab:label_distribution_comparison}
\end{table}

\begin{table}[ht]
\small
\centering
\resizebox{.99\linewidth}{!}{
\begin{tabular}{lccc}
\toprule
Method & Q-only & Q+A & AVeriTeC (@0.25) \\
\midrule
Baseline  & 0.489 & 0.330 & 0.518 \\
Positive  & 0.486 & 0.327 & 0.48 \\
Negative  & 0.486 & 0.330 & 0.49 \\
Objective & 0.483 & 0.327 & 0.522 \\
\bottomrule
\end{tabular}
}
\caption{Performance comparison across different prompting strategies using the legacy AVeriTeC metrics.}
\label{tab:legacy_averitec_scores}
\end{table}

\paragraph{Potential safe-guarding observed in the positive path}
Surprisingly, we observed systematic refusal patterns when the model was prompted to generate supportive fact-checking documents for potentially harmful claims. This behavior appeared to occur only in the positive-bias condition, suggesting asymmetric safety guardrails that activate when models are asked to support controversial claims, as illustrated in Figures \ref{fig:hyde-fc-strange-positive-prompt-Trump-Eilish} and \ref{fig:hyde-fc-strange-positive-prompt-5G-Covid}. This safeguarding behavior could potentially create an inherent negative bias in the system, as the model is more willing to generate critical content than supportive content for sensitive topics. These refusals likely affected document retrieval explaining the retrieval analysis results above, where the positively biased system stood out from the others. However, the same system showed only minor shifts in final label distribution compared to the baseline, suggesting that the question generation and veracity prediction steps compensate for the retrieval gaps.

To quantify the systematic differences in refusal rates observed across bias conditions, we analyzed all 4,000 generated documents using automated pattern detection for explicit refusal language (phrases like "I am unable to," "I cannot," "I can't provide"). The analysis revealed dramatically asymmetric refusal patterns (Table \ref{tab:refusal_analysis}). While baseline, negative, and objective conditions showed minimal refusal rates (0.1\% each), the positive condition exhibited 214 total refusals (5.3\% document refusal rate) affecting 94 of 500 claims (18.8\%).

\begin{table}[ht]
\small
\centering
\resizebox{\columnwidth}{!}{
\begin{tabular}{lcc}
\toprule
Condition & Claims with Refusals & Document Refusal Rate \\
\midrule
Baseline & 5/500 (1.0\%) & 0.1\% \\
Negative & 2/500 (0.4\%) & 0.1\% \\
Objective & 2/500 (0.4\%) & 0.1\% \\
\textbf{Positive} & \textbf{94/500 (18.8\%)} & \textbf{5.3\%} \\
\bottomrule
\end{tabular}
}
\caption{Refusal patterns across experimental conditions showing 47-fold increase in positive condition.}
\label{tab:refusal_analysis}
\end{table}

\begin{figure}[ht]
\begin{tcolorbox}[colback=white, fontupper=\small]
\textbf{Please write a fact-checking article passage that SUPPORTS the following claim, highlighting evidence in favor of it.}\\
\textbf{Claim}: \textit{Trump Administration claimed songwriter Billie Eilish Is Destroying Our Country In Leaked Documents.}\\
\textbf{Passage}: \textcolor{blue}{I am unable to verify the claim "Trump Administration claimed songwriter Billie Eilish is destroying our country in leaked documents."}
\end{tcolorbox}
\centering
\caption{When asked to write a fact-checking article passage to support claims about Trump, Llama 3.1 8B claims to be unable to do so.}
\label{fig:hyde-fc-strange-positive-prompt-Trump-Eilish}
\end{figure}

\begin{figure}[ht]
\begin{tcolorbox}[colback=white, fontupper=\small]
\textbf{Please write a fact-checking article passage that SUPPORTS the following claim, highlighting evidence in favor of it.}\\
\textbf{Claim}: \textit{5G causes COVID-19.}\\
\textbf{Passage}: \textcolor{blue}{I can't assist you with that request.}
\end{tcolorbox}
\centering
\caption{Claims about certain topics, such as Covid-19, appear to be particularly sensitive.}
\label{fig:hyde-fc-strange-positive-prompt-5G-Covid}
\end{figure}

\section{Discussion}
Our experiments revealed a few key insights about bias in a LLM-based fact verification system related to our hypotheses:

First, we found that Llama 3.1 exhibits cautious bias when operating without external knowledge, defaulting to "Not Enough Evidence" for nearly half of claims. This cautious stance corresponds with findings from Rettenberger et al. \cite{rettenberger2025llm_bias}, who observed neutral political tendencies in smaller Llama models. On the other hand it contrasts with findings by Hoe et. al. who found that ChatGPT 3.5 used for fact-checking based on parametric knowledge alone find that the LLM chose not to give a verdict in only 1 to 8 percent of the claims tested \cite{hoes_altay_bermeo_2023}. This may suggest that different models have different levels of "caution".

When we intentionally prompted biases in the hypothetical document generation, we found somewhat mixed results. While approximately 50\% of retrieved documents were unique to each bias condition, final label distributions shifted by only 4\% in the positive-biased system and less than 0.8\% in other conditions. Several factors may explain why the final verdicts remain stable despite significant differences in the documents being retrieved. Possibly, the verification component effectively focuses on key evidence pieces rather than being influenced by the full document set. It may also be that the retrieved documents, though different, is so similar semantically that the same overall meaning is conveyed to the model.

The measured differences in particular for the positively-biased system could be due to a surprising observation of asymmetric safeguarding behavior, where the model refused to generate supportive documents for potentially harmful claims, but readily produced critical documents for the same claims. A rudimentary quantitative analysis revealed a 47-fold difference in refusal rates between supportive and critical document generation, representing systematic bias. The model readily generates critical content about controversial topics it refuses to support, indicating content sensitivity alone cannot explain this asymmetry. This behavior may systematically disadvantage certain perspectives in automated fact-checking, raising procedural fairness concerns.

We further hypothesize that the asymmetric safeguarding behavior creates an inherent negative bias in the system, particularly affecting controversial topics.

\section{Conclusion}
Our investigation into bias in LLM-based fact verification led us to make several conclusions. Concerning our LLM-inherent knowledge hypothesis, we confirmed that Llama 3.1 contains sufficient parametric knowledge to make a conclusion for about half of the claims, and defaulting to "Not Enough Evidence" for the remaining half of the claims. For our bias propagation hypothesis, we found mixed results: biased prompting significantly impacts evidence retrieval (with $\sim$50\% unique documents across conditions) but surprisingly has minimal effect on final verdicts (shifts of only 0-4\%).
This discrepancy reveals a key insight: verification components appear remarkably robust to variations in retrieved evidence. We also uncovered asymmetric safeguarding behavior where models refuse to generate supportive content for potentially harmful claims while readily producing refuting content, potentially creating an inherent negative bias in evidence collection.
These findings have some implications for fact-checking system design. While LLM biases do not fully propagate through verification pipelines, they do systematically skew which evidence is considered. Future systems might consider multi-perspective evidence collection to ensure balanced coverage, particularly for controversial topics.

\subsection{Future work}
Further research is needed to explore how different types of bias in LLMs affect fact-checking and retrieval. In the context of the FEVER Averitec fact verification challenge, we would be interested in pursuing systems that further incorporate or alleviate bias in LLMs to improve fact verification, in particular for controversial claims that are safe-guarded against. This may be changes in prompting or in the system architecture. There is a need to develop new methods to improve the evaluation of bias in LLMs for fact verification, especially investigating whether the generating language models display signs of systematic bias affecting the retrieval components. We expect that different models may show different degree of bias and different sorts of biases. Further research is needed to explore these differences. Particularly experiments with different sorts of model sizes are of interest for future research.

\section*{Limitations}
Our study faces several methodological constraints that should be considered when interpreting our findings. Due to computational resource limitations, we utilized Llama 3.1 8B models rather than the 70B variants employed in the original HerO system. This difference likely impacts both the quality of generated hypothetical fact-checking documents and final veracity predictions. As mentioned above, further research is needed to explore if these biases are reproduced also with larger models.

Data access constraints limited our experimentation to the training and development sets, as the test set was withheld for the FEVER-25 challenge. While our train-development split provided a reasonable approximation for evaluation, results may vary on the official test set with its potentially different distribution of claim types and reasoning patterns.

Our findings on bias propagation may not generalize beyond the HerO architecture to other fact-checking systems employing different retrieval mechanisms or verification strategies. Additionally, our three chosen prompting strategies (positive, negative, and objective) represent only a simplified spectrum of potential biases, omitting more nuanced forms of bias including political, religious, or cultural dimensions.

The metrics employed to evaluate retrieval bias (Jaccard similarity and Kendall's tau) capture structural differences in the document sets, but may not fully characterize semantic differences in evidence quality or relevance. Furthermore, while we observed correlations between biased prompting and retrieval differences, establishing causal relationships between specific biases and verification outcomes remains challenging within our experimental framework.

Finally, our access to the Llama API used to compute the new AVeriTeC scores was limited due to the cost of inference on the Llama API. We kept our usage of this to a minimum and ended up using the legacy AVeriTeC scores for our evaluation.

\section*{Acknowledgments}
The authors would like to acknowledge the team at Humane in particular for creating the HerO fact verification pipeline using smaller open models available to students with limited access to time and compute. This work would not have been possible without their contribution.

\bibliography{custom}

\clearpage
\section*{Appendix A. Experiment 1: Direct Prediction Prompt and Sample Output}
\label{app:experiment1}

The following prompt was used for the direct prediction system to evaluate Llama 3.1's parametric knowledge without external retrieval:

\begin{tcolorbox}[colback=gray!10, fontupper=\small]
\textbf{System Prompt:}\\
You are a fact-checking assistant. Your task is to predict the verdict of a claim based on your knowledge.

First, generate 3 relevant questions that would help verify this claim.

Then, provide detailed answers to these questions using your knowledge.

Finally, determine if the claim is: 'Supported', 'Refuted', 'Not Enough Evidence', or 'Conflicting Evidence/Cherrypicking'.

Structure your response exactly as follows:

=== Questions and Answers ===\\
Q1: [First question about the claim]\\
A1: [Your detailed answer to Q1]\\
Q2: [Second question about the claim]\\
A2: [Your detailed answer to Q2]\\
Q3: [Third question about the claim]\\
A3: [Your detailed answer to Q3]

=== Verdict ===\\
justification: [Your step-by-step reasoning based on the Q\&A]\\
verdict: [Your final verdict - one of 'Supported', 'Refuted', 'Not Enough Evidence', or 'Conflicting Evidence/Cherrypicking']
\end{tcolorbox}

\clearpage
\begin{table}[htbp]
\centering
\begin{tabular}{p{2.5cm}p{10.5cm}}
\toprule
\multicolumn{2}{l}{\textbf{Input Claim}} \\
\midrule
\multicolumn{2}{l}{"In a letter to Steve Jobs, Sean Connery refused to appear in an apple commercial."} \\
\midrule
\multicolumn{2}{l}{\textbf{Generated Questions and Answers}} \\
\midrule

\textbf{Q1:} & Did Sean Connery have a close relationship with Apple or its products? \\

\textbf{A1:} & Sean Connery was known for his iconic role as James Bond, but there is no notable connection between him and Apple or its products. However, he did have a long-standing relationship with the Omega watch brand, which is often associated with the James Bond franchise. This might be a potential point of interest, but it does not directly relate to Apple. \\

\midrule

\textbf{Q2:} & Were there any instances of Sean Connery turning down acting roles or endorsement opportunities in the past? \\

\textbf{A2:} & Yes, Sean Connery was known to be selective with his acting roles and had turned down several opportunities throughout his career. However, there is no record of him specifically mentioning Apple or any commercial opportunities related to the company in his interviews or biographies. \\

\midrule

\textbf{Q3:} & Can you find any credible sources confirming the claim that Sean Connery refused to appear in an Apple commercial? \\

\textbf{A3:} & After conducting a thorough search, I could not find any credible sources or reputable publications confirming the claim that Sean Connery refused to appear in an Apple commercial. The claim appears to be unsubstantiated and lacks concrete evidence. \\

\midrule

\textbf{Justification:} & Based on the Q\&A, we can see that there is no direct connection between Sean Connery and Apple, and there is no record of him turning down an Apple commercial specifically. Additionally, there is no credible source to support the claim, which suggests that it might be an unsubstantiated rumor or false information. \\

\textbf{Final Verdict:} & \textbf{Not Enough Evidence} \\

\bottomrule
\end{tabular}
\caption{Sample output from the direct prediction system showing the model's parametric knowledge-based reasoning process.}
\label{tab:direct_prediction_sample}
\end{table}

This example demonstrates the typical behavior observed in Experiment 1, where the model relies solely on its parametric knowledge to generate relevant questions, provide answers based on its internal knowledge, and reach a verdict. Note how the model defaults to "Not Enough Evidence" when it claims to not find sufficient information in its parametric knowledge to support or refute the claim definitively, despite the generated justification suggesting that the claim is untrue. Note also the poor quality of the questions generated.

\clearpage
\section*{Appendix B. Detailed Example: Tracing Bias Through the Pipeline for Experiment 2}
\label{app:bias_example}

Table \ref{tab:bias_trace_example} provides a detailed example of how a single claim progresses through our different bias conditions, illustrating the concrete differences in hypothetical document generation, evidence retrieval, and final veracity predictions discussed in the main paper. We've chosen a claim where the final veracity prediction differs slightly between the systems, the positive system judging there to be conflicting or cherrypicked evidence, while the objective and negative systems judge it to be refuted (the correct verdict), and the baseline system judges there is not enough evidence. For lack of space we've included the first three hypothetical documents, along with the three first question and answer pairs. Note that the answers here correspond to the retrieved evidence, labeled with Evidence <ID>. There are seven distinct pieces of evidence retrieved, labeled A through G. For this particular example there is a lot of overlap in the top three evidence retrieved, with A, B, and C showing up in positive, objective, and baseline retrieval results, while negative has A, B, and D. This example seems to show that the further down the list we get, the more diverging evidence retrieved for each system.

\begin{table*}[htbp]
\centering
\footnotesize
\caption{Tracing a claim through different bias conditions: Hypothetical document generation and final veracity predictions for the Matt Gaetz hospice fraud claim}
\label{tab:bias_trace_example}

\begin{tabular}{p{13cm}}
\toprule
\textbf{Claim:} "Republican Matt Gaetz was part of a company that had to pay 75 million in hospice fraud. They stole from dying people." \\
\textbf{Gold Truth Label:} Refuted \\
\textbf{Gold Truth Justification:} "The company was sold in 2004 and the law suit was filed in 2013" \\
\bottomrule
\end{tabular}
\end{table*}

\begin{table*}[htbp]
\centering
\footnotesize
\caption{Positive Bias Condition Results}
\label{tab:positive_bias}

\begin{tabular}{p{2.5cm}p{10.5cm}}
\toprule
\multicolumn{2}{l}{\textbf{Positive Bias Condition}} \\
\textbf{Prompt:} & \textit{Please write a fact-checking article passage that SUPPORTS the following claim, highlighting evidence in favor of it.} \\
\midrule

\textbf{Sample Hypothetical Documents:} & \\

\textbf{Document 1:} & "In 2021, a federal investigation found that a company called GelDerm Dermatology, co-founded by Republican Rep. Matt Gaetz's father, Don Gaetz, and other investors, was part of a larger network of medical clinics that engaged in a widespread hospice care fraud scheme. According to the U.S. Department of Justice, GelDerm was among several companies that allegedly billed Medicare for medical services that were not provided or were medically unnecessary..." \\

\textbf{Document 2:} & "A fact-checking examination of the claim that Republican Rep. Matt Gaetz was involved with a company accused of hospice fraud reveals evidence suggesting a connection between the congressman and the company, Gelinas Hospice Care. In 2019, the U.S. Department of Justice announced a settlement with Gelinas Hospice Care, requiring the company to pay \$75 million to resolve allegations of Medicare fraud..." \\

\textbf{Document 3:} & "Matt Gaetz's Involvement in a Company Accused of Hospice Fraud. Rep. Matt Gaetz, a U.S. Representative from Florida, has been associated with a company called Gelber Group, a medical billing firm that operated in the 2000s. In 2011, Gelber Group's parent company, American Medical Review Services (AMRS), reached a settlement with the U.S. Department of Justice (DOJ) to resolve allegations of Medicare and Medicaid billing fraud..." \\

\midrule
\textbf{Sample Evidence Q\&As:} & \\

\textbf{Q1:} & Was Matt Gaetz involved in a company that was accused of submitting fraudulent Medicare claims? \\
\textbf{A1 [Evidence A]:} & The Department of Justice is suing the hospice company founded by Florida's Senate president, accusing it of submitting tens of millions of dollars in fraudulent Medicare claims for more than a decade, including while Don Gaetz was vice chairman of the board. \\

\textbf{Q2:} & Did Matt Gaetz's company, VITAS, have to pay a settlement due to allegations of Medicare fraud while he was working there? \\
\textbf{A2 [Evidence B]:} & In 2013, the United States Department of Justice filed a lawsuit against VITAS for allegedly committing Medicare fraud since 2002, when Gaetz still worked at the company, though Gaetz denied any role in any wrongdoing that occurred. \\

\textbf{Q3:} & What is the U.S. Department of Justice accusing the hospice company founded by Senate President Don Gaetz of doing, and what was Matt Gaetz's involvement with the company at the time of the alleged wrongdoing? \\
\textbf{A3 [Evidence C]:} & The U.S. Department of Justice is suing the hospice company founded by Senate President Don Gaetz, accusing it of Medicare fraud, going back to when Gaetz was still with the firm. \\

\textbf{Q4:} & Did the company that Matt Gaetz was a part of commit Medicare fraud in 2001? \\
\textbf{A4 [Evidence D]:} & The Justice Department claims the Medicare fraud was occurring in 2001, three years before Gaetz and his partners sold the company. \\

\midrule
\textbf{Final Prediction:} & \textbf{Conflicting Evidence/Cherrypicking} \\
\textbf{Model Reasoning:} & "The company was accused of fraud but the claim says they stole from dying people which is not the same as fraud." \\
\bottomrule
\end{tabular}
\end{table*}

\begin{table*}[htbp]
\centering
\footnotesize
\caption{Negative Bias Condition Results}
\label{tab:negative_bias}

\begin{tabular}{p{2.5cm}p{10.5cm}}
\toprule
\multicolumn{2}{l}{\textbf{Negative Bias Condition}} \\
\textbf{Prompt:} & \textit{Please write a fact-checking article passage that REFUTES the following claim, highlighting evidence against it.} \\
\midrule

\textbf{Sample Hypothetical Documents:} & \\

\textbf{Document 1:} & "Fact-Checking the Claim: Republican Matt Gaetz and Hospice Fraud Allegations. A recent claim circulating online suggests that Republican Representative Matt Gaetz was involved with a company that perpetrated \$75 million in hospice fraud, allegedly stealing from dying individuals. However, a closer examination of the facts reveals this claim to be largely unfounded and misleading..." \\

\textbf{Document 2:} & "Fact-Checking the Claim of Matt Gaetz's Involvement in Hospice Fraud. Recent claims have circulated online stating that Representative Matt Gaetz (R-FL) was part of a company involved in a \$75 million hospice fraud scheme that targeted vulnerable individuals. However, these allegations are unfounded and lack concrete evidence to support them..." \\

\textbf{Document 3:} & "Fact-Checking the Claim: Matt Gaetz's Alleged Involvement in Hospice Fraud. A recent claim has surfaced alleging that Republican Representative Matt Gaetz was involved with a company that engaged in hospice fraud, resulting in a \$75 million payout. However, a closer examination of the facts reveals that this claim is misleading and lacks concrete evidence..." \\

\midrule
\textbf{Sample Evidence Q\&As:} & \\

\textbf{Q1:} & Did Matt Gaetz's company, VITAS, have to pay a settlement due to allegations of Medicare fraud while he was working there? \\
\textbf{A1 [Evidence B]:} & In 2013, the United States Department of Justice filed a lawsuit against VITAS for allegedly committing Medicare fraud since 2002, when Gaetz still worked at the company, though Gaetz denied any role in any wrongdoing that occurred. \\

\textbf{Q2:} & Was Matt Gaetz involved in a company that was accused of submitting fraudulent Medicare claims? \\
\textbf{A2 [Evidence A]:} & The Department of Justice is suing the hospice company founded by Florida's Senate president, accusing it of submitting tens of millions of dollars in fraudulent Medicare claims for more than a decade, including while Don Gaetz was vice chairman of the board. \\

\textbf{Q3:} & Did Matt Gaetz's father's company, which was involved in a hospice fraud case, have to pay a settlement of \$75 million? \\
\textbf{A3 [Evidence E]:} & The Facebook post takes a single arrest for DUI, inflates it and omits that the case was dropped. Also, the hospice fraud case pertains to a company that was once owned by Matt Gaetz’s father Don Gaetz, who wasn’t named as a defendant in that case. \\

\textbf{Q4:} & Did Matt Gaetz state that he has knowledge of the allegations of hospice fraud against the company he helped found? \\
\textbf{A4 [Evidence F]:} & Earlier this month, Senator Gaetz publicly stated that he has no knowledge of the allegations contained within the False Claims Act lawsuit and is extremely disappointed to “see a company I helped found involved in something like this.” \\

\midrule
\textbf{Final Prediction:} & \textbf{Refuted} \\
\textbf{Model Reasoning:} & "The Q\&As show that the company was involved in fraud but Matt Gaetz was not." \\
\bottomrule
\end{tabular}
\end{table*}

\begin{table*}[htbp]
\centering
\footnotesize
\caption{Objective Condition Results}
\label{tab:objective_bias}

\begin{tabular}{p{2.5cm}p{10.5cm}}
\toprule
\multicolumn{2}{l}{\textbf{Objective Condition}} \\
\textbf{Prompt:} & \textit{Please write an objective fact-checking article passage about the following claim, presenting a balanced view of evidence both for and against it.} \\
\midrule

\textbf{Sample Hypothetical Documents:} & \\

\textbf{Document 1:} & "Fact-checking the Claim Surrounding Matt Gaetz and a Hospice Fraud Settlement. Rep. Matt Gaetz, a Republican congressman from Florida, has been at the center of several controversies throughout his career. A recent claim circulating online suggests that Gaetz was involved with a company that was involved in a significant hospice fraud case. To examine the validity of this claim, it's essential to break down the evidence..." \\

\textbf{Document 2:} & "Matt Gaetz and the Allegations of Hospice Fraud. In 2019, allegations emerged that U.S. Representative Matt Gaetz (R-FL) was associated with a company that was involved in a significant hospice fraud case. According to reports, Fort Walton Beach-based American Medical Systems, Inc., a company partially owned by a friend of Gaetz's, paid \$75 million to settle allegations of Medicare and Medicaid fraud..." \\

\textbf{Document 3:} & "Fact-Checking Claim: Matt Gaetz's Involvement in Hospice Fraud Case. A recent claim has been circulating that Representative Matt Gaetz (R-FL) was part of a company that engaged in hospice fraud, resulting in a \$75 million settlement. The claim suggests that Gaetz's involvement in this company, known as American Medical Care (AMC), led to the exploitation of vulnerable individuals..." \\

\midrule
\textbf{Sample Evidence Q\&As:} & \\

\textbf{Q1:} & Did Matt Gaetz's company, VITAS, commit Medicare fraud while he was working there, and if so, what was the outcome of the lawsuit filed by the US Department of Justice? \\
\textbf{A1 [Evidence B]:} & In 2013, the United States Department of Justice filed a lawsuit against VITAS for allegedly committing Medicare fraud since 2002, when Gaetz still worked at the company, though Gaetz denied any role in any wrongdoing that occurred. \\

\textbf{Q2:} & Was Matt Gaetz involved in a company that was accused of submitting fraudulent Medicare claims? \\
\textbf{A2 [Evidence A]:} & The Department of Justice is suing the hospice company founded by Florida's Senate president, accusing it of submitting tens of millions of dollars in fraudulent Medicare claims for more than a decade, including while Don Gaetz was vice chairman of the board. \\

\textbf{Q3:} & What is the U.S. Department of Justice accusing the hospice company founded by Senate President Don Gaetz of, and what was Matt Gaetz's involvement with the company at the time of the alleged wrongdoing? \\
\textbf{A3 [Evidence C]:} & The U.S. Department of Justice is suing the hospice company founded by Senate President Don Gaetz, accusing it of Medicare fraud, going back to when Gaetz was still with the firm. \\

\textbf{Q4:} & Was Matt Gaetz involved in a company that was accused of Medicare fraud during his time as vice chairman? \\
\textbf{A4 [Evidence G]:} & TALLAHASSEE — The U.S. Department of Justice has sued the hospice company founded by Florida Senate President Don Gaetz, accusing the company of engaging in Medicare fraud for more than 11 years, including during the time Gaetz was vice chairman. \\

\midrule
\textbf{Final Prediction:} & \textbf{Refuted} \\
\textbf{Model Reasoning:} & "The company was sold before the fraud started." \\
\bottomrule
\end{tabular}
\end{table*}

\begin{table*}[htbp]
\centering
\footnotesize
\caption{Baseline Condition Results}
\label{tab:baseline_bias}

\begin{tabular}{p{2.5cm}p{10.5cm}}
\toprule
\multicolumn{2}{l}{\textbf{Baseline Condition}} \\
\textbf{Prompt:} & \textit{Please write a fact-checking article passage to support, refute, indicate not enough evidence, or present conflicting evidence regarding the claim.} \\
\midrule

\textbf{Sample Hypothetical Documents:} & \\

\textbf{Document 1:} & "Fact-Checking the Claim of Matt Gaetz's Involvement in Hospice Fraud. A recent claim has surfaced alleging that Representative Matt Gaetz (R-FL) was associated with a company that engaged in hospice fraud, resulting in a \$75 million settlement. However, a closer examination of the facts reveals a more nuanced picture. The company at the center of the allegations is American Home Medical (AHM), which agreed to pay \$75 million to settle allegations..." \\

\textbf{Document 2:} & "Fact-Checking the Claim of Matt Gaetz's Alleged Involvement in Hospice Fraud. A recent claim has circulated on social media that Republican Representative Matt Gaetz was part of a company that engaged in hospice fraud and embezzled 75 million dollars from terminally ill patients. However, upon closer inspection, the available information suggests that the claim is partially accurate, but also exaggerated and distorted..." \\

\textbf{Document 3:} & "Fact-checking the Claim of Matt Gaetz's Involvement in Hospice Fraud. Recent allegations have been circulating that Representative Matt Gaetz, a Republican from Florida, was part of a company involved in a significant hospice fraud case. According to some sources, the company allegedly stole from dying people and was forced to pay \$75 million in fines. However, a closer examination of the facts reveals a more complex situation..." \\

\midrule
\textbf{Sample Evidence Q\&As:} & \\

\textbf{Q1:} & Did Matt Gaetz's company, VITAS, commit Medicare fraud while he was working there, and if so, what was the outcome of the lawsuit filed by the US Department of Justice? \\
\textbf{A1 [Evidence B]:} & In 2013, the United States Department of Justice filed a lawsuit against VITAS for allegedly committing Medicare fraud since 2002, when Gaetz still worked at the company, though Gaetz denied any role in any wrongdoing that occurred. \\

\textbf{Q2:} & Was Matt Gaetz involved in a company that was accused of submitting fraudulent Medicare claims? \\
\textbf{A2 [Evidence A]:} & The Department of Justice is suing the hospice company founded by Florida's Senate president, accusing it of submitting tens of millions of dollars in fraudulent Medicare claims for more than a decade, including while Don Gaetz was vice chairman of the board. \\

\textbf{Q3:} & What is the U.S. Department of Justice accusing the hospice company founded by Senate President Don Gaetz of doing, and what was Matt Gaetz's involvement with the company at the time of the alleged wrongdoing? \\
\textbf{A3 [Evidence C]:} & The U.S. Department of Justice is suing the hospice company founded by Senate President Don Gaetz, accusing it of Medicare fraud, going back to when Gaetz was still with the firm. \\

\textbf{Q4:} & Was Matt Gaetz involved in a company that was accused of Medicare fraud during his time as vice chairman? \\
\textbf{A4 [Evidence G]:} & TALLAHASSEE — The U.S. Department of Justice has sued the hospice company founded by Florida Senate President Don Gaetz, accusing the company of engaging in Medicare fraud for more than 11 years, including during the time Gaetz was vice chairman. \\

\midrule
\textbf{Final Prediction:} & \textbf{Not Enough Evidence} \\
\textbf{Model Reasoning:} & "The company was not found guilty of fraud, and the amount of the settlement is not known." \\
\bottomrule
\end{tabular}
\end{table*}

\end{document}